\documentclass[9pt,twocolumn]{extarticle}
\usepackage{authblk}
\usepackage{cuted}
\usepackage[margin=1.9cm,columnsep=0.6cm]{geometry}
\usepackage{times}
\usepackage{amsmath,amssymb}
\usepackage{graphicx}
\usepackage{booktabs}
\usepackage{caption}
\usepackage[numbers,sort&compress]{natbib}
\usepackage{titlesec}
\usepackage{abstract}
\usepackage{float}
\usepackage{color}
\usepackage{xcolor}
\usepackage{hyperref}
\usepackage{stfloats}

\titlespacing*{\section}{0pt}{8pt}{4pt}
\titlespacing*{\subsection}{0pt}{6pt}{3pt}

\title{\vspace{-1.5em} TT-net: Quantum Inspired Tensor Network Denoising
in Conditional GANs\vspace{-0.5em}}
\author{Michal A. Sterzel$^*$, Marko J. Rančić$^\dagger$ \\
University of Luxembourg, FSTM, Department of Computer Science
Campus Belval,  6 Av. de la Fonte, L-4364 Esch-Belval, Esch-sur-Alzette, Luxembourg\\
\texttt{$^*$michal.sterzel@gmail.com}   
\texttt{$^\dagger$marko.rancic@uni.lu}}

\date{}

\begin{document}
\twocolumn[\begin{@twocolumnfalse}
\maketitle
\begin{abstract}
Developed as a workhorse for classical simulations of quantum algorithms and quantum many-body systems, Tensor Network methods have entered the scientific mainstream in quantum physics. Among various types of tensor networks, Tensor Trains (commonly know as Matrix Product States in the quantum computing community) have already found applications in machine learning. These methods often rely on a powerful linear algebra tool called the Singular Value Decomposition (SVD).
Several conditional GAN architectures for image denoising incorporate
SVD as a single-cut decomposition step
applied to generator feature maps. In this work we introduce TT-Net, which replaces the
per-channel SVD denoising block with a two-cut tensor-train
decomposition capable of accessing cross-channel information directly, a
capability absent from contemporary alternatives. In a controlled
comparison differing only in this decomposition mechanism, TT-Net
outperforms SVD-Net on PSNR and SSIM across all three noise types tested
(Gaussian, motion blur, and salt-and-pepper), supporting the hypothesis that
cross-channel access improves denoising quality. Training-dynamics analysis
further shows that TT-Net's adversarial loss term consistently saturates to
a stagnant state across all three noise types, more so than SVD-Net's,
while reconstruction quality continues to improve regardless, raising an
open question about the adversarial component's contribution that this
work identifies but does not resolve. Furthermore, for Gaussian noise our method outperforms both the EigenGAN and the state of the art Pix2pix method which does not assume any linear algebra decompositions and does not retain any linear algebra information. Our manuscript shows how quantum inspired tools can be used as practical real world feature filters for deep learning applications. 
\end{abstract}
\vspace{1em}
\end{@twocolumnfalse}]

\section{Introduction}

\label{sec:intro}
Quantum computing is a novel computational paradigm deemed to bring the next disruption in the world of computing. As hardware matures the hunt for a ''killer application" is at full swing. The desire to simulate larger and larger quantum circuit has given birth to the fields of Tensor Networks, with most commonly used subtype of these networks being called Tensor Trains, commonly referred to as Matrix Product States in quantum physics literature~\cite{orus2014tensornetworks}. 

The task of recovering a clean image from a degraded version known 
as image denoising is a long-standing problem in image processing, with
applications ranging from medical imaging and remote sensing to consumer
photography, in each case improving the reliability of downstream data analysis.
Generative adversarial networks (GANs)~\cite{goodfellow2014gan} have been
widely adopted for this task in their conditional form, where a generator is
trained to transform a degraded image to its clean counterpart while a
discriminator provides adversarial feedback, typically achieving better
preservation of fine detail than classical denoising methods that rely on
fixed assumptions about noise or image structure.

Within this space, several architectures incorporate Singular Value
Decomposition (SVD) as a component of the generator, using it in different
roles: adjusting a generated feature map's singular values toward a target's,
as in EigenGAN~\cite{eigengan}, or applying it as a per-channel filtering
mechanism, as in SVD-Net~\cite{svdnet}. In every such case, however, SVD is
applied as a single decomposition step: the feature map is unfolded into one
matrix and decomposed once. Since a convolutional feature map is naturally a
three-axis object (channels, height, and width) a single SVD cut can only
ever separate one such grouping of axes (e.g.\ channels from combined spatial
dimensions) from the rest, and whichever axes are merged in that unfolding can no
longer be examined separately.

This raises a natural question: does generalizing a single SVD cut into a
sequence of cuts, known as tensor-train decomposition, change what a denoising mechanism of this kind can capture or discard? Tensor decomposition methods have
previously been applied inside neural architectures, including
GANs~\cite{tgan,ttnn}, but exclusively to compress learned weight matrices,
never to filter feature-map activations on a per-sample basis, and never in
a denoising context (Section~\ref{sec:related}).

This paper introduces \textbf{TT-Net} a method inspired in Matrix Product States, replacing SVD-Net's per-channel
SVD denoising block ~\cite{svdnet} with a two-cut tensor-train decomposition, inserted at
the same locations in the generator and evaluated under identical training
conditions, isolating the decomposition mechanism as the sole variable of
interest. Our contributions are: (1) a tensor-train denoising block that,
unlike SVD-Net, allows the decomposition to access cross-channel structure
directly; (2) an adaptive per-cut thresholding scheme that preserves a
comparable overall retained-energy target to SVD-Net, enabling a fair,
like-for-like comparison; and (3) an experimental comparison of TT-Net
against SVD-Net, EigenGAN, and a non-SVD Pix2pix baseline across three
distinct noise types.

\section{Related Work}
\label{sec:related}

Several conditional GAN architectures for image denoising incorporate SVD in
different roles. EigenGAN~\cite{eigengan} applies SVD within the generator's
decoder, adjusting the singular values of a generated feature map toward
those of the target image and including their difference as an explicit loss
term, making its use of SVD supervised and target-aware. SVD-Net~\cite{svdnet},
the direct predecessor of this work, instead applies SVD purely as a
filtering mechanism: after each encoder downsampling stage, every
feature-map channel is decomposed independently via SVD, and only the
singular values needed to retain a fixed fraction of that channel's energy
are kept, with no comparison to a target and no dedicated loss term.

Tensor decomposition methods have separately been applied to compress the
parameters of generative and general neural network architectures.
\citet{tgan} replace fully-connected layers of a GAN with
Tucker-decomposition-based tensor layers, achieving substantial parameter
compression with limited effect on sample quality. \citet{ttnn} similarly
represent fully-connected layer weight matrices in tensor-train format,
reporting compression factors of several orders of magnitude. Both approaches
decompose learned weight matrices once, with the resulting cores trained
end-to-end. Neither applies a tensor decomposition to feature-map activations
on a per-sample basis, and neither is evaluated in a denoising context.

This leaves a gap: no existing work replaces a single-cut, SVD-based
activation filter of the kind used in SVD-Net with a multi-cut tensor-train
decomposition applied in the same architectural role. TT-Net, introduced
next, addresses this gap directly.

\section{Method}
\label{sec:method}

\subsection{SVD-Net Recap}
SVD-Net's denoising block decomposes each channel of a feature map independently: for a
feature map of shape channels $\times$ height $\times$ width, SVD is applied separately
to each channel's own height $\times$ width matrix, retaining enough singular values to
preserve a fixed fraction $\theta=0.9$ of that channel's energy before reconstruction.
Because no two channels ever appear in the same matrix, this mechanism cannot compare
channels to one another or exploit any relationship between them, regardless of the
threshold chosen.

\subsection{TT-Net Architecture}
TT-Net replaces SVD-Net's per-channel SVD with a two-cut tensor-train decomposition,
inserted at the same three locations in the generator (after each of the three encoder
downsampling stages) (Table~\ref{tab:shapes}) and reconstructing a filtered feature map that is passed forward
exactly as SVD-Net's would be.

\begin{table}[t]
\centering
\small
\begin{tabular}{@{}lccc@{}}
\toprule
Insertion point & Channels & Height & Width \\
\midrule
After 1st encoder stage & 64 & 32 & 32 \\
After 2nd encoder stage & 128 & 16 & 16 \\
After 3rd encoder stage & 256 & 8 & 8 \\
\bottomrule
\end{tabular}
\caption{Feature map shape at each TT-Net / SVD-Net insertion point.}
\label{tab:shapes}
\end{table}

For a single image's feature map $\mathcal{X} \in \mathbb{R}^{C \times H \times W}$, the
first cut reshapes $\mathcal{X}$ into $M_1 \in \mathbb{R}^{C \times (HW)}$, keeping
channel as its own axis while merging height and width. SVD is applied directly to
$M_1 = U_1 \Sigma_1 V_1^T$, so that all channels of a given image are rows of the same
matrix and are compared against one another during decomposition, a capability
absent from SVD-Net. The top $\chi_1$ components by retained energy (Section
\ref{sec:threshold}) form $\text{Core}_1 = U_1[:,1{:}\chi_1]$. The corresponding
remainder $R_1 = \Sigma_1[1{:}\chi_1,1{:}\chi_1]\,V_1^T[1{:}\chi_1,:]$ is carried forward
rather than reconstructed immediately, while any components beyond $\chi_1$ are
discarded entirely.

$R_1$ is reshaped so that channel merges with height, giving
$M_2 \in \mathbb{R}^{(\chi_1 H) \times W}$, and a second SVD $M_2 = U_2 \Sigma_2 V_2^T$
yields an adaptively chosen $\chi_2$, giving $\text{Core}_2 \in
\mathbb{R}^{\chi_1 \times H \times \chi_2}$ (reshaped from $U_2[:,1{:}\chi_2]$) and
$\text{Core}_3 = \Sigma_2[1{:}\chi_2,1{:}\chi_2]\,V_2^T[1{:}\chi_2,:] \in
\mathbb{R}^{\chi_2 \times W}$. The filtered feature map is reconstructed by contracting
all three cores:
\[
\widetilde{\mathcal{X}}[i,j,k] = \sum_{a=1}^{\chi_1} \sum_{b=1}^{\chi_2}
\text{Core}_1[i,a]\,\text{Core}_2[a,j,b]\,\text{Core}_3[b,k].
\]
All other components of the architecture such as the generator/discriminator structure, loss
functions, optimizer, and training procedures are identical to SVD-Net, isolating the
decomposition mechanism as the sole variable under study.

\subsection{Adaptive Threshold for Fair Comparison}
\label{sec:threshold}
Both models select their retained rank adaptively per image, as the smallest $\chi$ for
which cumulative squared-singular-value energy reaches a threshold $\theta$, rather than
fixing $\chi$ in advance. Because TT-Net's second cut operates on the remainder passed
forward from the first, its two thresholds compound multiplicatively rather than
additively: setting $\theta_1=\theta_2=0.9$ would yield an overall retained energy closer
to $0.9^2 = 0.81$, not $0.9$. To keep the comparison to SVD-Net's single $\theta=0.9$
target fair, TT-Net's thresholds are instead set to $\theta_1 = \theta_2 = \sqrt{0.9}
\approx 0.9487$, so that $\theta_1\times\theta_2\approx0.9$. Since $\chi$ is chosen as
the smallest sufficient value and singular values are discrete, actual retained energy
typically exceeds the nominal threshold.

\subsection{Cross-Channel Information Gained}
The two cuts extract structurally different information. The first cut's dominant
direction (largest singular value) corresponds to the strongest pattern of agreement
across channels, while smaller-singular-value directions capture channel-specific,
residual variation. Retaining only the dominant directions keeps what channels share
while discarding what is unique to individual channels, information a
per-channel decomposition cannot access at all. The second cut, operating on what the
first retained with width as its only free axis, instead captures horizontal spatial
regularities (e.g.\ smooth gradients, repeated textures, edges) shared jointly across the
retained channel directions and height positions. Whether discarding what is excluded at
either cut removes noise rather than genuine detail is the empirical question addressed
in Section~\ref{sec:results}.

\section{Experimental Setup}
\label{sec:setup}

All experiments use the CIFAR-10 dataset, resized to $64\times64$ pixels and
split into 40{,}000 training, 10{,}000 validation, and 10{,}000 test images.
Three synthetic noise types are applied to the datasets to generate noisy-clean pairs for supervised training, applied identically across all four models: additive Gaussian
noise with standard deviation $0.5$, motion blur with an $11$-pixel
directional kernel at $45^\circ$, and salt-and-pepper noise affecting $20\%$
of pixels.

All models are trained for $50$ epochs. Evaluation uses Peak Signal-to-Noise
Ratio (PSNR) and Structural Similarity Index (SSIM)~\citep{psnr_ssim} on the held-out test
set, computed identically across all four models. SSIM is computed using 
the standard windowed formulation (11x11 Gaussian window, sigma=1.5), 
consistent with the original definition, computed identically across all four models.

\subsection{Optimized Hyperparameters}
\label{sec:hyperparams}

\begin{table}[t]
\centering
\small
\begin{tabular}{@{}lcccc@{}}
\toprule
Model & lr (G, D) & $\beta_1$ & $\beta_2$ & Optimizer \\
\midrule
Pix2pix  & $2\times10^{-4}$ & $0.5$ & $0.999$ & Adam\\
EigenGAN & $2\times10^{-3}$ & $0.5$ & $0.999$ & Adam\\
SVD-Net  & $1\times10^{-4}$ & $0.9$ & $0.999$ & Adam\\
TT-Net   & $1\times10^{-4}$ & $0.9$ & $0.999$ & Adam\\
\bottomrule
\end{tabular}
\caption{Optimized hyperparameters used for each model, with a uniform batch
size of 32 applied across all four.}
\label{tab:hyperparams}
\end{table}

Table~\ref{tab:hyperparams} lists the hyperparameters used across training for all models. 
A single hyperparameter was intentionally unified across all four models despite this per-model tuning: batch size, fixed at 32 for every model and every noise type. Keeping batch size fixed at 32, while preserving each model's literature-recommended learning rate and momentum, isolates batch size as the only hyperparameter standardized across the entire comparison.

Each model's optimizer settings follow the configuration reported in its own
originating paper: Pix2pix and EigenGAN's values come directly from
\citet{pix2pix} and \citet{eigengan} respectively, and SVD-Net's from
\citet{svdnet}. TT-Net has no independent originating paper of its own to
draw from, since it is introduced in this work; rather than selecting a new,
untested configuration, TT-Net instead adopts SVD-Net's exact optimizer
settings unchanged. This choice follows directly from TT-Net's design as a
direct architectural extension of SVD-Net (Section~\ref{sec:method}),
differing only in its denoising block: using SVD-Net's own recommended
configuration keeps that comparison a single-variable ablation, isolating
the decomposition mechanism as the only difference between the two models,
rather than introducing an additional, independently-tuned hyperparameter
difference alongside it.

\subsection{Implementation details}
\label{sec:impl_details}
Beyond the stated hyperparameters the four models also differ architecturally in discriminator design and reconstruction loss. SVD-Net and TT-Net share an unconditional discriminator trained with cross-entropy loss, and a generator with a reconstruction loss blending MSE and MS-SSIM ($\delta=0.84$). Concretely, for a generated image $\hat{Y}$ and its clean target $Y$, the two reconstruction terms are the pixel-wise mean squared error
\[
\mathcal{L}_{L2} = \|\hat{Y} - Y\|_2^2,
\]
and the multi-scale structural similarity loss
\[
\mathcal{L}_{\text{MS-SSIM}} = 1 - \text{MS-SSIM}(\hat{Y}, Y),
\]
which are combined into the reconstruction loss
\[
\mathcal{L}_{\text{recon}} = \delta \cdot \mathcal{L}_{\text{MS-SSIM}} + (1-\delta) \cdot \mathcal{L}_{L2}, \quad \delta = 0.84.
\]
The generator's adversarial loss term, $\mathcal{L}_{G_{adv}}$, is the standard non-saturating GAN generator loss. The full generator objective is then
\[
\mathcal{L}_G = \mathcal{L}_{G_{adv}} + \mathcal{L}_{\text{recon}}.
\]
Pix2pix and EigenGAN instead share a conditional
PatchGAN discriminator with label-smoothed binary cross-entropy and an
L1-based reconstruction loss generator component, following their own original architectures.
One deviation from the original SVD-Net architecture is disclosed for
transparency: an additional downscaling stage was introduced in the encoder
to reduce computation and fit within available GPU job-time constraints.
This modification is inherited unchanged by TT-Net, preserving the validity
of the SVD-Net vs. TT-Net comparison, though it does mean SVD-Net/TT-Net use
a shallower encoder bottleneck than Pix2pix/EigenGAN, a further reason the
latter two are treated as contextual baselines rather than controlled
comparison points.

\section{Results}
\label{sec:results}

\subsection{Quantitative Results}

\begin{table*}[t]
\centering
\small
\begin{tabular}{@{}lcccccc@{}}
\toprule
 & \multicolumn{2}{c}{Gaussian} & \multicolumn{2}{c}{Motion Blur} & \multicolumn{2}{c}{Salt \& Pepper} \\
\cmidrule(lr){2-3}\cmidrule(lr){4-5}\cmidrule(lr){6-7}
Model & PSNR & SSIM & PSNR & SSIM & PSNR & SSIM \\
\midrule
Pix2pix  & 25.52 & 0.7466 & \textbf{36.18} & \textbf{0.9643} & \textbf{41.79} & \textbf{0.9881} \\
EigenGAN & 25.18 & 0.7545 & 26.71 & 0.8398 & 34.24 & 0.9513 \\
SVD-Net  & 23.60 & 0.7455 & 18.02$^\dagger$ & 0.4260$^\dagger$ & 28.98 & 0.8761 \\
TT-Net   & \textbf{26.28} & \textbf{0.7968} & 26.34 & 0.8058 & 29.97 & 0.8926 \\
\bottomrule
\end{tabular}
\caption{Test-set PSNR and SSIM (standard windowed formulation) across noise
types, using each model's optimized hyperparameters (Section~\ref{sec:hyperparams})
with a uniform batch size of 32. Bold indicates the best value across \emph{all
four} models. TT-Net achieves the best PSNR and SSIM of any model on Gaussian
noise. $^\dagger$SVD-Net's motion-blur result reflects a late-training
discriminator collapse rather than a converged model (Section~\ref{sec:stability}).}
\label{tab:results}
\end{table*}

Table~\ref{tab:results} lists all PSNR and SSIM values achieved on the test image set for all four models across the three noise types. Within the controlled comparison
(Section~\ref{sec:setup}), TT-Net outperforms a common single channel alternative SVD-Net on every metric and
every noise type: by $2.68$ dB PSNR and $0.0513$ SSIM on Gaussian noise, by
$8.32$ dB PSNR and $0.3798$ SSIM on motion blur, and by $0.99$ dB PSNR and
$0.0165$ SSIM on salt-and-pepper noise. This consistent advantage supports
the hypothesis introduced in Section~\ref{sec:intro}: allowing the
denoising mechanism to access cross-channel structure, rather than treating
every channel in isolation, corresponds to a measurable improvement in
reconstruction quality across all three noise conditions tested.

The motion-blur gap is by far the largest of the three; however, it should be
interpreted with care: as discussed in Section~\ref{sec:stability},
SVD-Net's motion-blur training exhibits a sustained discriminator-driven
collapse partway through training that TT-Net's run does not, meaning part
of this specific gap likely reflects a training-stability difference rather
than the denoising mechanism alone. The Gaussian and salt-and-pepper gaps,
where no comparable instability was observed in either run, are the more
direct evidence of the mechanism's effect.

Pix2pix and EigenGAN, included as context from a different architecture
family achieve higher
absolute scores on motion blur and salt-and-pepper than either SVD-Net or
TT-Net, plausibly reflecting their deeper four-stage encoder bottleneck in
addition to any effect of their denoising mechanism (or lack thereof).
Since architecture, discriminator design, and hyperparameters all differ
simultaneously between these two models and the SVD-Net/TT-Net pair, this
comparison does not isolate the contribution of SVD in the same controlled
way and is not the focus of this paper's analysis.

Beyond the SVD-Net comparison, TT-Net also achieves the best PSNR and SSIM of any of the four models on Gaussian noise (Table~\ref{tab:results}), outperforming Pix2pix by 0.76 dB PSNR and 0.0502 SSIM, and EigenGAN by 1.10 dB PSNR and 0.0423 SSIM. This is notable because Pix2pix and EigenGAN's deeper four-stage encoder (Section~\ref{sec:impl_details}) generally advantages them on the other two noise types; TT-Net's lead here cannot be attributed to that architectural advantage on account of SVD-Net performing worse than either Pix2pix and EigenGAN, and more directly reflects the cross-channel decomposition mechanism itself.

\subsection{Qualitative Results}

\begin{figure*}[t]
\centering
\includegraphics[width=\textwidth]{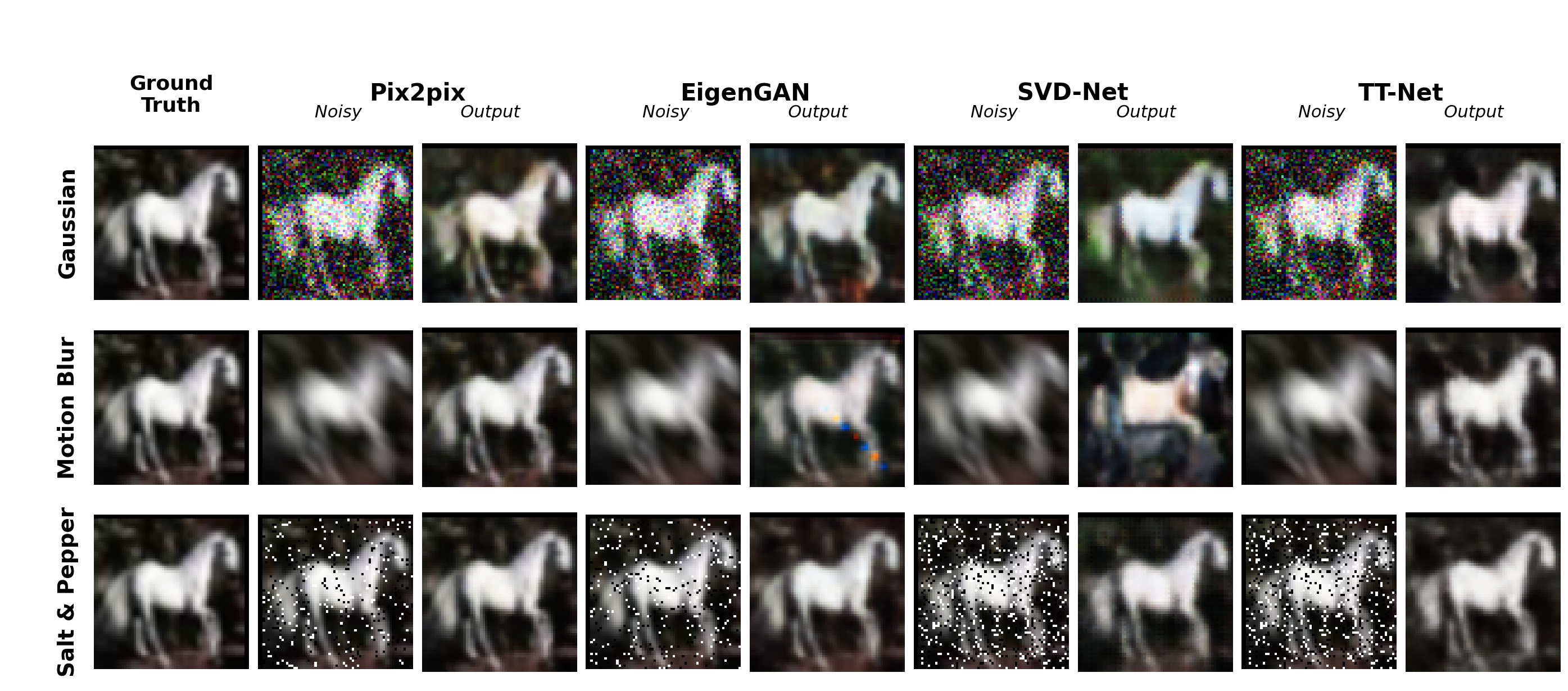}
\caption{Representative test-set outputs across all four models and three
noise types. Ground truth is shown once per row since it is identical
across models. Each model's own noisy input and output are shown alongside
it.}
\label{fig:qualitative}
\end{figure*}

Figure~\ref{fig:qualitative} shows representative test-set examples for one
CIFAR-10 image per noise type, with all four models' noisy inputs and
outputs shown alongside the shared ground truth. Only the noise realization 
differs between them, since noise is generated independently per run.

Visual differences between SVD-Net and TT-Net are easily visible for the Gaussian noise. 
TT-Net's output preserves the ground truth's natural coloring, whereas SVD-Net's output is noticeably saturated with a green tint. On motion blur the gap is stark and immediate: SVD-Net's output shows a clearly corrupted patch inconsistent with the underlying image structure, a direct visual symptom of the discriminator collapse discussed in Section 5.3, while TT-Net's output remains a coherent, if blurry, reconstruction. On salt-and-pepper noise, outputs from the two models are visually similar, consistent with the smaller quantitative gap in Table~\ref{tab:results}.

Pix2pix and EigenGAN's outputs appear visually sharper on motion blur and
salt-and-pepper, consistent with their higher quantitative scores on those
noise types, though as discussed in Section~\ref{sec:setup} this cannot be
attributed to their denoising mechanism alone given the architectural and
hyperparameter differences involved.

\subsection{Training Stability}
\label{sec:stability}

SVD-Net's motion-blur run exhibits a training failure distinct from every
other run in this study. Training proceeds normally for the first six
epochs, reaching a peak of PSNR $22.13$ / SSIM $0.5984$ at epoch $6$, with
$D(\text{real})$ and $D(\text{fake})$ close to the balanced point of $0.5$.
From epoch $7$ onward, $D(\text{real})$ climbs steadily and almost
monotonically to $0.876$ by epoch $50$, while $D(\text{fake})$ falls to
$0.106$ over the same span (Figure~\ref{fig:app-discprobe}). PSNR and SSIM collapse alongside this shift and
never recover, ending at PSNR $18.02$ / SSIM $0.4260$ (Figure~\ref{fig:app-psnr}, Figure~\ref{fig:app-ssim}). This is consistent
with a discriminator that overpowers the generator partway through
training, rather than a stable convergence to a lower-quality result.

Every other run in this study, for both SVD-Net and TT-Net, instead
converges toward the opposite extreme: $D(\text{real})$ and $D(\text{fake})$
settle to almost exactly $0.500$ within the first few epochs,
indicating a discriminator no longer able to distinguish real from
generated images. This is not unique to TT-Net: SVD-Net 
shows a smaller-magnitude version of this pattern on Gaussian and salt-and-pepper noise, settling near $0.500$ within the first 2-5 epochs, followed on Gaussian by a gradual drift from epoch 21 and a sharper shift at epoch 39, after which D(real) settles near 0.59-0.60 for the remainder of training, a smaller-scale version of the motion-blur failure. TT-Net shows the initial saturation pattern on all three noise types, settling fully within the first two epochs for Gaussian, by epoch $10$ for salt-and-pepper, and by epoch $14$ for motion blur (Figure~\ref{fig:app-discprobe}).

TT-Net's adversarial loss term ($G_{adv}$) declines rapidly early in training before settling to a stable minimum for the remainder of training across all three noise types (Figure~\ref{fig:app-genloss}). For motion blur and salt-and-pepper, this minimum is essentially constant for the rest of training, while on Gaussian it instead fluctuates without a clear trend around a similar level. This is in clear contrast to SVD-Net's $G_{adv}$, which continues to move measurably throughout training, most visibly on Gaussian and motion blur (Figure~\ref{fig:app-genloss}). The reconstruction-based loss terms (L2, MS-SSIM) do not saturate in the same way for either model, continuing to change throughout training (Figure~\ref{fig:app-genloss}, Figure~\ref{fig:app-msssim}). On Gaussian both models noticeably fluctuate across epochs, however TT-Net has a decreasing trend as opposed to SVD-Net. On motion blur and salt-and-pepper, TT-Net's MS-SSIM loss changes smoothly with visibly less epoch-to-epoch fluctuation than SVD-Net's. Specifically on motion blur, TT-Net's MS-SSIM loss decreases smoothly and monotonically for the remainder of training, while SVD-Net's fluctuates without a clear trend around a substantially worse, largely unimproving level, consistent with the discriminator-driven collapse described above.

In every case where a run settles toward this $0.5/0.5$ equilibrium, PSNR and SSIM continue to improve for the remainder of training rather than stalling, with the best validation result reached at or near the final epoch (Figure~\ref{fig:app-psnr}, Figure~\ref{fig:app-ssim}). This is visible in the total generator loss as well: despite $G_{adv}$ itself having flattened, TT-Net's total loss keeps declining throughout training, driven almost entirely by the still-improving reconstruction term rather than by adversarial feedback (Figure~\ref{fig:app-total-loss}). This suggests that once training reaches this state, continued improvement is driven by the reconstruction loss operating through each model's denoising mechanism rather than by adversarial feedback, since a discriminator outputting a constant $0.5$ regardless of input provides no informative gradient to the generator. However, confirming that this is specifically attributable to the implementation of the cross-channel mechanism in TT-Net as described in Section~\ref{sec:method} would require training TT-Net with the adversarial loss removed entirely, which we identify as a direction for future work in Section~\ref{sec:conclusion}.

\section{Conclusion and Future Work}
\label{sec:conclusion}

This paper introduced TT-Net, which replaces SVD-Net's per-channel SVD
denoising block with a two-cut tensor-train decomposition capable of
accessing cross-channel structure, and compared it directly against
SVD-Net in a controlled setting differing only in this mechanism. TT-Net
outperformed SVD-Net in every metric and every noise type tested (Gaussian,
motion blur, salt-and-pepper), supporting the hypothesis that allowing a
denoising mechanism to compare channels against each other, rather than
treating each in isolation, corresponds to a measurable improvement in
reconstruction quality. TT-Net also achieved the best overall PSNR and SSIM of all four models tested on Gaussian noise, ahead of Pix2pix and EigenGAN despite their deeper encoder.

Training-dynamics analysis further showed that TT-Net's adversarial loss
term consistently reaches a stagnant state across all three
noise types, more completely and consistently than SVD-Net's, while
reconstruction quality continues to improve throughout training regardless.
This raises the question of whether the adversarial component contributes
meaningfully to TT-Net's performance at all, a question this work
identifies but does not resolve.

Several limitations bound these conclusions. The motion-blur comparison is
confounded by a distinct training failure in SVD-Net's run, unrelated to
the denoising mechanism itself.  Reported test metrics use each model's 
final-epoch checkpoint rather than its best validation checkpoint.

Future work includes directly testing whether TT-Net's adversarial loss term is
necessary at all, by training with it removed entirely and comparing
results to those reported here. Extending the comparison to additional datasets, and analyzing whether TT-Net's per-block retained rank correlates with denoising difficulty
across noise types are additional natural extensions of this work.

\appendix
\section{Supporting Training Curves}
\label{sec:appendix}

\begin{figure*}[b]
\centering
\includegraphics[width=\textwidth]{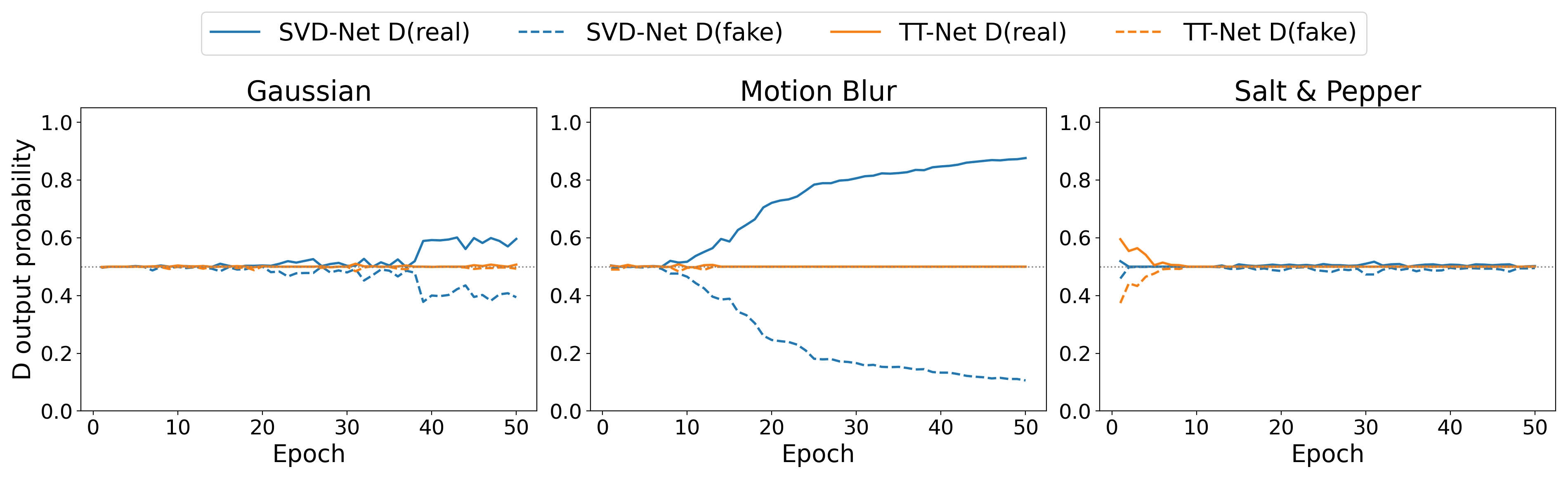}
\caption{Discriminator output probabilities $D(\text{real})$/$D(\text{fake})$
over training, SVD-Net vs. TT-Net, by noise type.}
\label{fig:app-discprobe}
\end{figure*}

\begin{figure*}[t]
\centering
\includegraphics[width=0.85\textwidth]{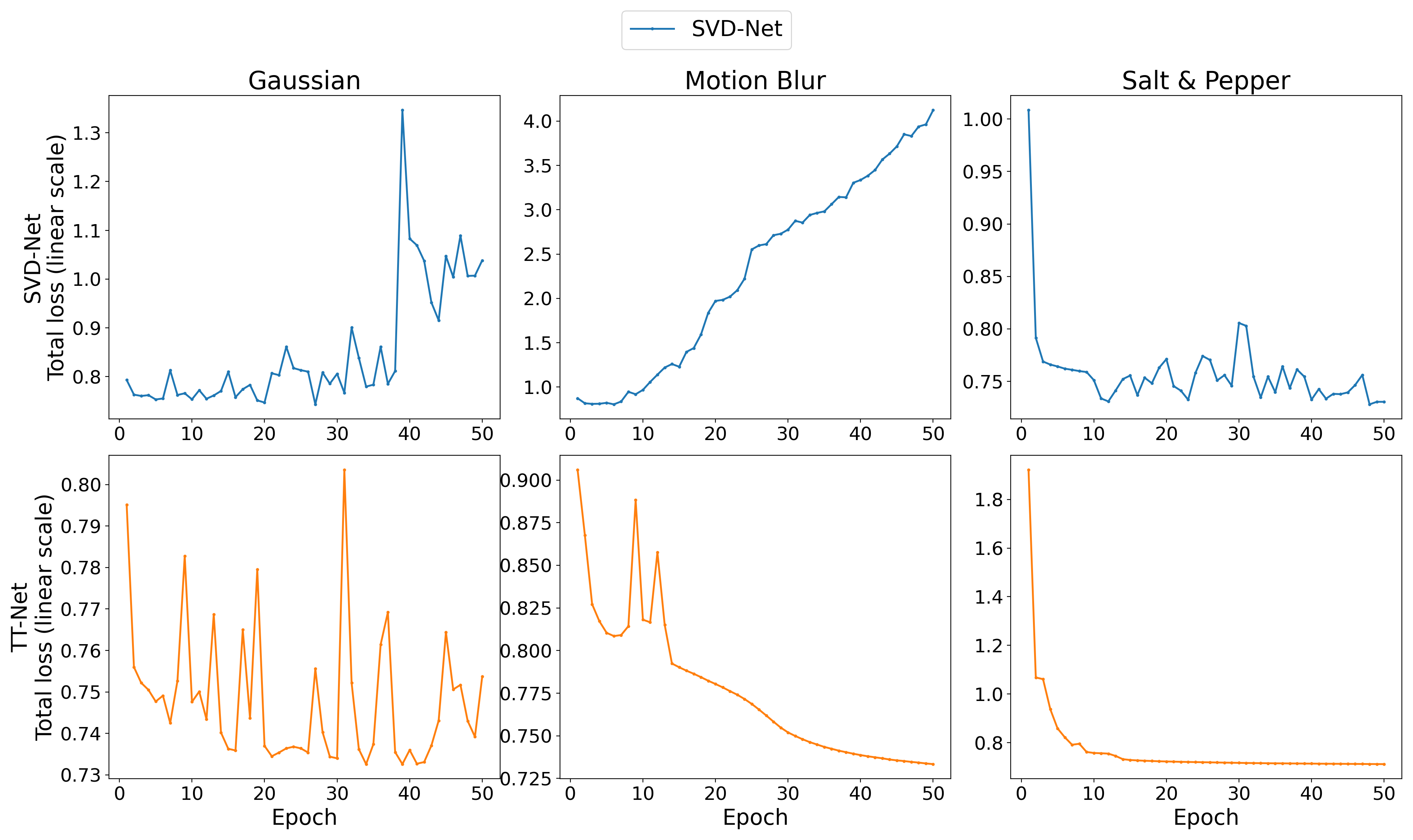}
\caption{Total generator loss ($\mathcal{L}_G = \mathcal{L}_{G_{adv}} + \mathcal{L}_{\text{recon}}$,
Section~\ref{sec:impl_details}) over training, SVD-Net (top) vs. TT-Net
(bottom), by noise type, linear scale.}
\label{fig:app-total-loss}
\end{figure*}

\begin{figure*}[t]
\centering
\includegraphics[width=0.85\textwidth]{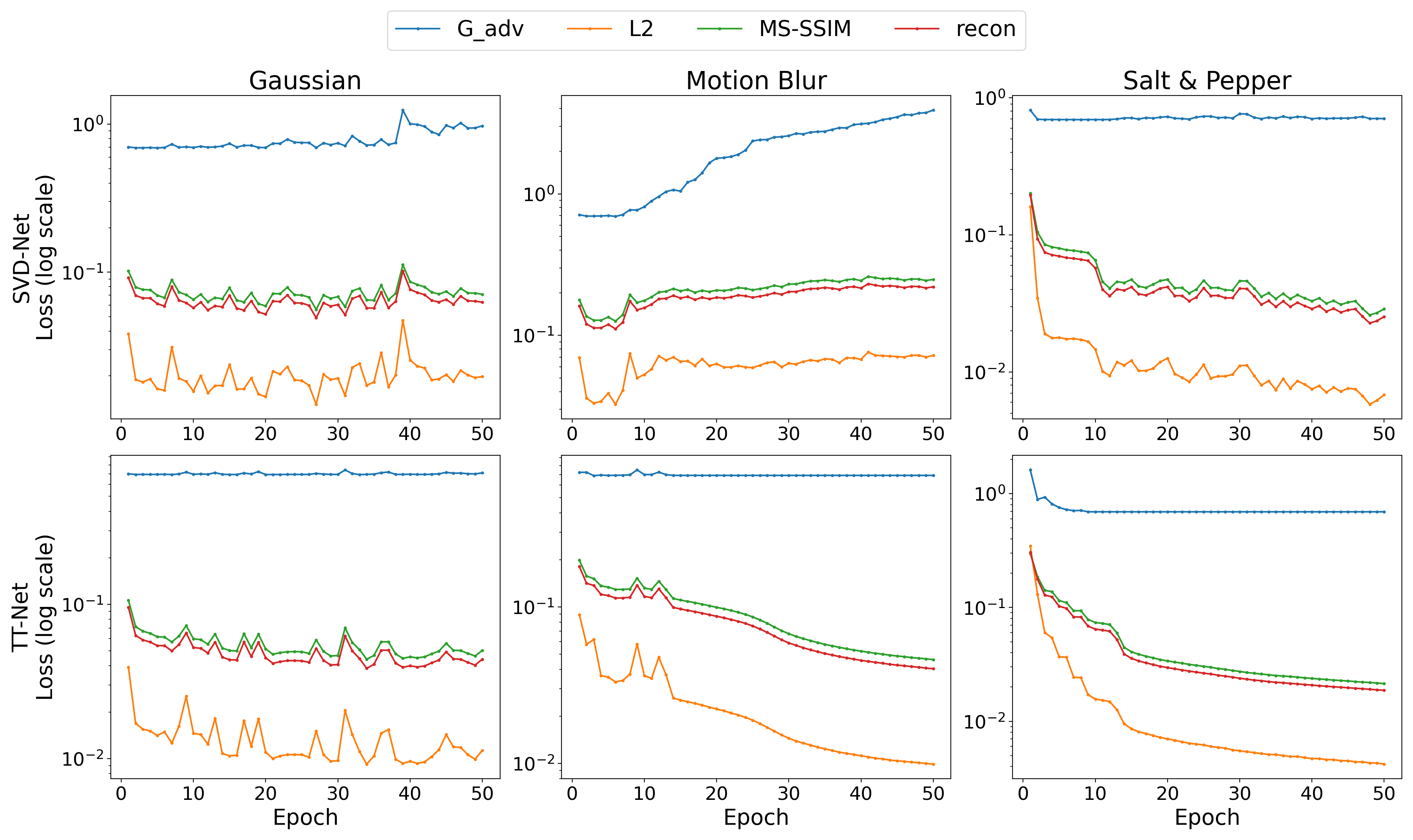}
\caption{Generator loss components ($G_{adv}$, L2, MS-SSIM, recon) over
training, SVD-Net (top) vs. TT-Net (bottom), by noise type. See
Section~\ref{sec:impl_details} for the loss definitions, $\mathcal{L}_{\text{recon}} = \delta \cdot \mathcal{L}_{\text{MS-SSIM}} + (1-\delta) \cdot \mathcal{L}_{L2},\, \delta = 0.84$.}
\label{fig:app-genloss}
\end{figure*}

\begin{figure*}[t]
\centering
\includegraphics[width=\textwidth]{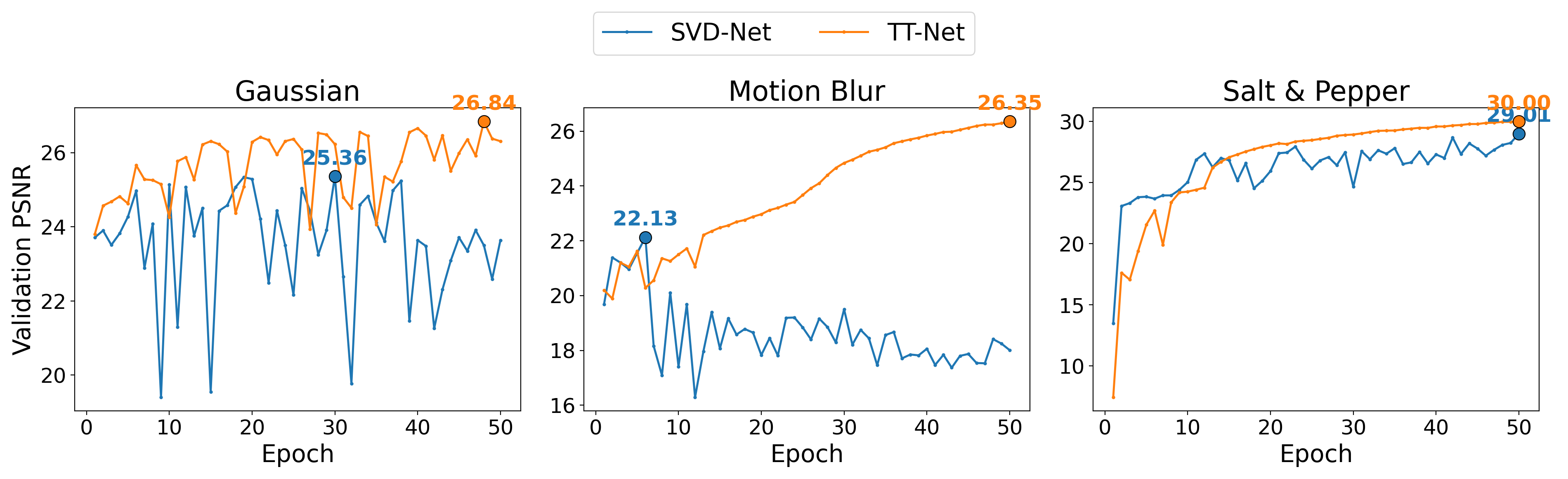}
\caption{Validation PSNR over training, SVD-Net vs. TT-Net, by noise type.}
\label{fig:app-psnr}
\end{figure*}

\begin{figure*}[t]
\centering
\includegraphics[width=\textwidth]{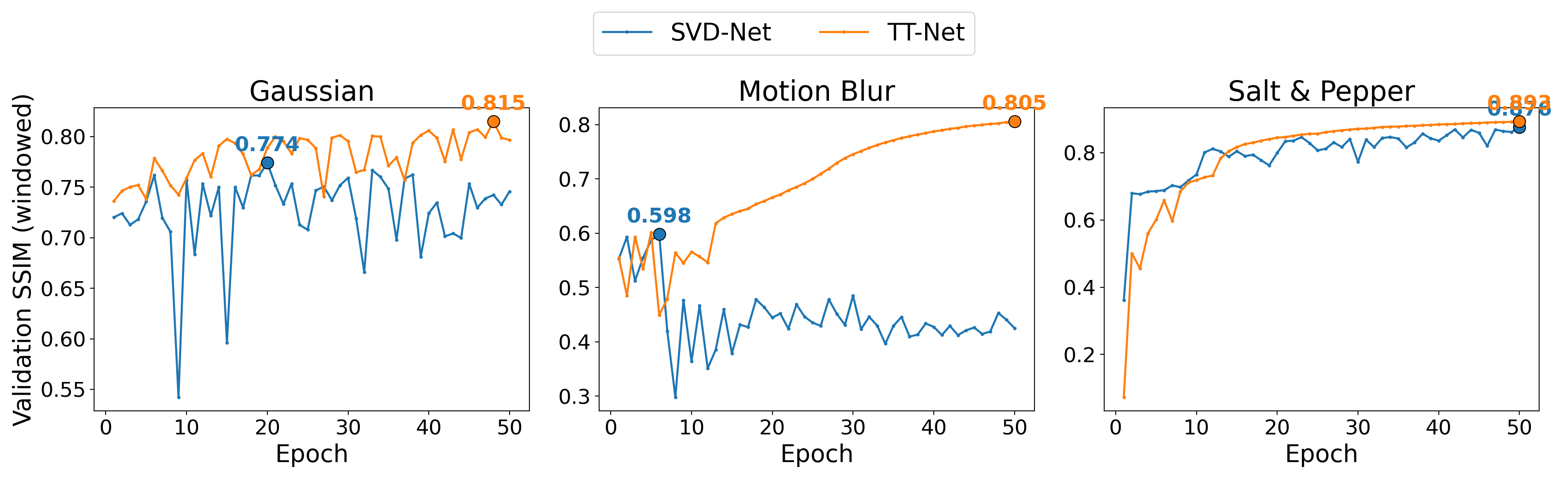}
\caption{Validation SSIM (windowed) over training, SVD-Net vs. TT-Net, by
noise type.}
\label{fig:app-ssim}
\end{figure*}

\begin{figure*}[t]
\centering
\includegraphics[width=\textwidth]{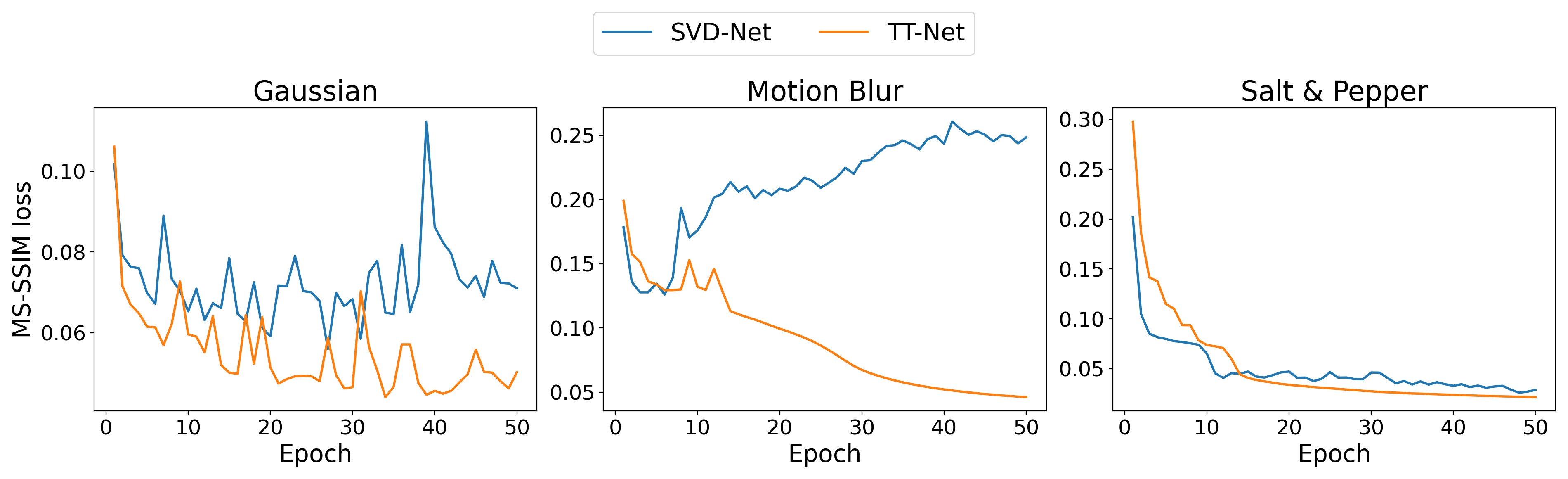}
\caption{MS-SSIM loss component over training, SVD-Net vs. TT-Net, by noise
type.}
\label{fig:app-msssim}
\end{figure*}

Figures~\ref{fig:app-discprobe}--\ref{fig:app-msssim} provide the
underlying training curves referenced in Section~\ref{sec:stability}, for
SVD-Net and TT-Net across all three noise types.

\newpage
\bibliographystyle{unsrtnat}
\bibliography{refs}

\end{document}